\documentclass[runningheads]{llncs}
\usepackage[T1]{fontenc}
\usepackage{amsmath}
\usepackage{amssymb}
\usepackage{multirow}
\usepackage{booktabs}
\usepackage{color}
\usepackage{silence}
\usepackage{graphicx}
\begin{document}
\title{Neural Fields for Continuous Periodic Motion Estimation in 4D Cardiovascular Imaging}

%
\titlerunning{INRs 4D Continuous Periodic Motion Estimation in Cardiac Data}
%
 
 
 

\author{Simone Garzia\inst{1,2} \and
Patryk Rygiel\inst{3} \and
Sven Dummer\inst{3} \and
Filippo Cademartiri\inst{4} \and
Simona Celi \inst{1} \and
Jelmer M. Wolterink \inst{3}}
\authorrunning{S. Garzia et al.}
%
\institute{BioCardioLab, Bioengineering Unit, Fondazione Monasterio, Massa 54100, Italy \and
Department of Information Engineering, University of Pisa, Pisa 56126, Italy \and Department of Applied Mathematics \& Technical Medical Centre, University of Twente, Enschede 7500 AE, The Netherlands \and Clinical Imaging Department, Fondazione Monasterio, Massa 54100, Italy
}

\maketitle 

\section*{Abstract}
Time-resolved three-dimensional flow MRI (4D flow MRI) provides a unique non-invasive solution to visualize and quantify hemodynamics in blood vessels such as the aortic arch. However, most current analysis methods for arterial 4D flow MRI use static artery walls because of the difficulty in obtaining a full cycle segmentation. To overcome this limitation, we propose a neural fields-based method that directly estimates continuous periodic wall deformations throughout the cardiac cycle.
For a 3D + time imaging dataset, we optimize an implicit neural representation (INR) that represents a time-dependent velocity vector field (VVF).
An ODE solver is used to integrate the VVF into a deformation vector field (DVF), that can deform images, segmentation masks, or meshes over time, thereby visualizing and quantifying local wall motion patterns. To properly reflect the periodic nature of 3D + time cardiovascular data, we impose periodicity in two ways. First, by periodically encoding the time input to the INR, and hence VVF. Second, by regularizing the DVF. We demonstrate the effectiveness of this approach on synthetic data with different periodic patterns,  ECG-gated CT, and 4D flow MRI data. The obtained method could be used to improve 4D flow MRI analysis. 


\section{Introduction}
Time-resolved three-dimensional flow MRI (4D flow MRI) provides unique insights into patient-specific cardiovascular conditions \cite{celi2017multimodality,berhane2020fully}. Visualization, quantification, and interpretation of 4D flow MRI require post-processing, which is typically performed under the assumption that the artery walls are static~\cite{frydrychowicz2011aortic,vali2019semi,berhane2020fully,garzia2023coupling}. However, for reliable determination of hemodynamic parameters such as wall shear stress, the wall deformation over the cardiac cycle should be considered \cite{capellini2021novel,calo2023impact,capellini2018computational}. Accurate tracking of the artery wall position remains a major challenge in 4D flow MRI imaging. Previous works have aimed to obtain time-dependent deformation fields based on automatic segmentation in different phases of the cardiac cycle, e.g.,  \cite{bustamante2023automatic,lian2023mri}. However, these techniques are limited by the resolution of the spatial grid and thus struggle to provide deformation fields that reflect the continuous nature of artery wall motion. As an alternative, deep learning-based registration approaches have been proposed, in which a neural network is trained to provide a deformation field, given cardiac imaging frames~\cite{ahn2020unsupervised,ahn2023co,ta2020semi}. However, such methods rely on the availability of large quantities of training data. 
In this work, we propose a one-shot image registration approach for motion estimation that does not require training data. Following recent developments in image registration~\cite{wolterink2022implicit,van2023robust}, we employ implicit neural representations (INRs)~\cite{sitzmann2020implicit}, also known as neural fields or coordinate networks. INRs represent a signal as a continuous function by associating to the input coordinates the value of an estimated function \cite{mildenhall2021nerf,wink2006phase}. Considering the target motion as a nonlinear, continuous, time-dependent periodic process, we represent a \textit{time-dependent and periodic} velocity vector field (VVF) with a trainable INR \cite{sun2022medical,han2023diffeomorphic,van2024deformable}. The obtained VVF is embedded in an ordinary differential equation (ODE) to estimate the final deformation vector field (DVF), which is trained with a single 4D image and can be applied to images and meshes representing anatomical structures such as the aortic wall. To reflect the periodic nature of cardiovascular data and to ensure periodicity of the time-dependent velocity field, we periodically embed the time input to the INR., Moreover, to impose periodicity on the obtained deformation fields, we propose a novel approach for periodic regularization. We assess the model's effectiveness using synthetic and real data sets and find that the proposed method provides continuous and periodic motion estimation.

\section{Methods}



\subsection{Temporal Diffeomorphic Image Registration}
To perform motion estimation, we jointly register multiple time frames within a period \(T\). Instead of estimating the displacement field $\phi_{0 \rightarrow 1}$ between a moving image ($I_0$) and a fixed one ($I_1$), we estimate the displacement fields $\phi_{t \to T}$ for a series of images $I_t$ over \(N\) time points $\{t_i\}_{i=0}^{N-1}$ with $t_0 = 0$ and $t_{N-1} = T$:
\begin{equation}
    \min_{\{\phi_{t_i \to T}\}_{i=0}^{N-2}} \sum_{i=0}^{N-2} \left[ \lVert I_{t_i} \circ \phi_{t_i \to T} - I_{T}  \rVert_2^2 \right],
    \label{eq:first_loss}
\end{equation}
We parameterize $\phi_{t_i \to T}$ via a general $\phi_t$. 
One approach for representing $\phi_t$ is the large deformation framework, 
such as the LDDMM method \cite{beg2005computing}. This approach breaks down the whole deformation process into multiple small steps by parameterizing a diffeomorphic deformation using an ODE:
\begin{equation}
    \frac{\partial \phi_t(\hat{P})}{\partial t} = v(\phi_t(\hat{P}), t), \, \phi_0(\hat{P}) = \hat{P}, \, \phi_{t \rightarrow T}:=\phi_T \circ \phi_t^{-1}
    \label{eq:ode}
\end{equation}
This iterative approach allows the transformations to be invertible and continuous, ensuring a global one-to-one mapping of each point in the two volumes \cite{ashburner2007fast,han2023diffeomorphic}. As the $\phi_t$ perfectly fits with the temporal nature of our data and because its invertibility fits the inductive bias that all the aortic data is diffeomorphic, we implemented a diffeomorphic registration framework using an ODE and a neural network parameterized VVF $v$. 

\begin{figure}[t!]
\includegraphics[width=\textwidth]{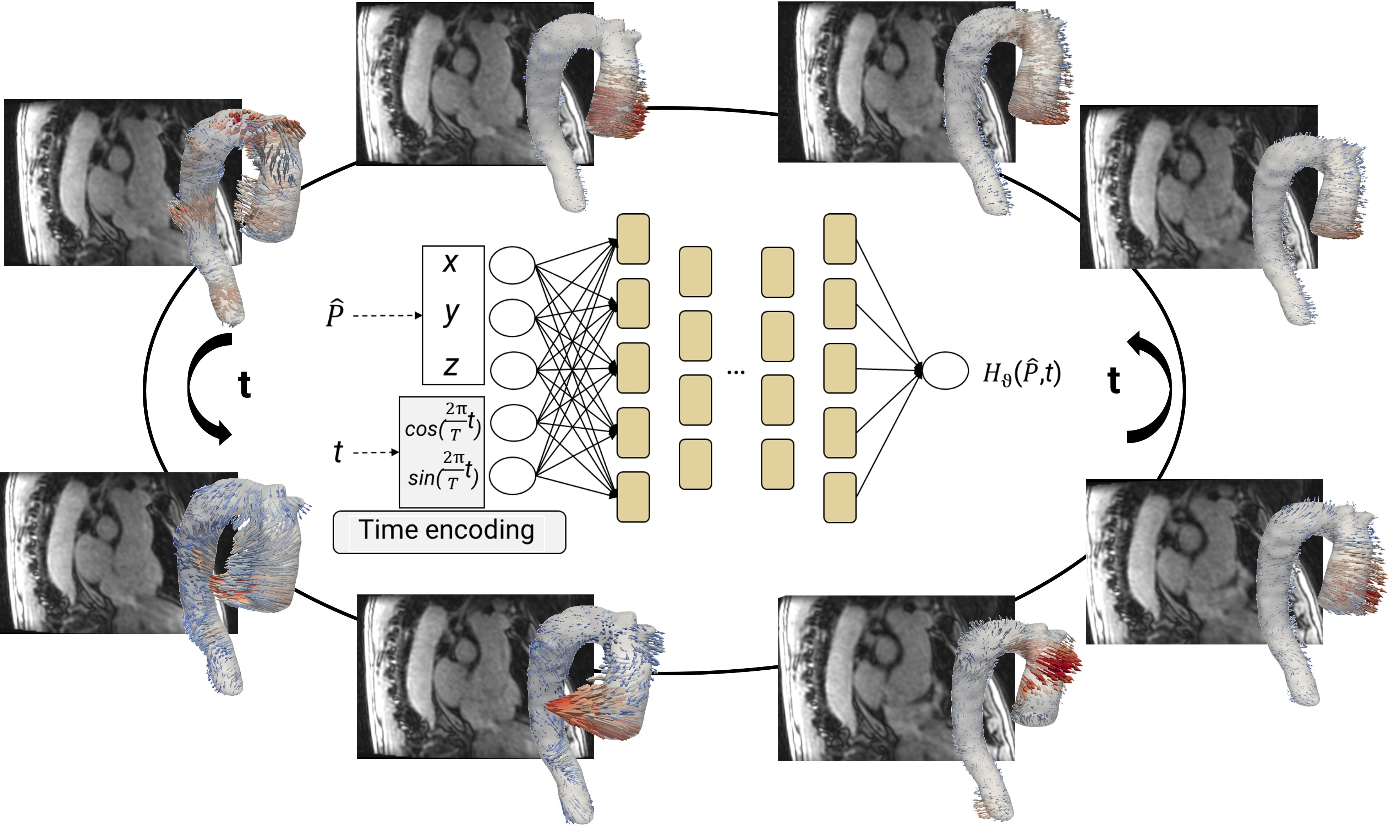}
\caption{Overview of the proposed method. We train an implicit neural representation $H_{\theta}\colon \mathrm{\Omega}\times [0, T] \to \mathbb{R}^3$ with spatial domain $\Omega \subset \mathbb{R}^3$, time horizon $T$, and weights $\theta$ to optimize. For every spatial point $\hat{P} = (x,y,z) \in \mathbb{R}^3$ and time frame $t$, $H_{\theta}(\hat{P}, t)$ represents the time-dependent velocity field describing the periodic motion at any point in space and time. We obtain a DVF by embedding the VVF $H_\theta$ in an ODE. Applying the DVF to the first time frame mesh allows for a full-cycle deformation.} 
\label{general_workflow}
\end{figure}


\subsection{Implicit Neural Representations}
The implicit neural representation that we use to estimate the time-dependent VVF is a neural network that maps a spatiotemporal coordinate to a vector. In particular, an INR representing a time-dependent VVF is a neural network $H_{\theta}\colon \mathrm{\Omega}\times [0, T] \to \mathbb{R}^3$ with spatial domain $\Omega \subset \mathbb{R}^3$, time horizon $T$, and weights $\theta$. Hence, given a spatial position $\hat{P}$ and time frame $t_{i}$, $H_{\theta}(\hat{P}, t_{i})$ is the velocity at $\hat{P}$ at time $t_{i}$. By parameterizing the velocity field with a function, we utilize spatial and temporal continuous modeling over multiple timeframes to accurately estimate the continuous DVF in Equation \eqref{eq:ode}.

\subsection{Periodic Velocity Fields}
While the standard INR parameterization of the VVF is time-dependent, it does not necessarily consider the periodicity in our cardiovascular imaging data. To ensure our network accurately represents specific time-dependent periodic VVFs, we use an INR $\Tilde{H}_\theta \colon \Omega \times \mathbb{R}^2 \to \mathbb{R}$ and let $H_\theta(\hat{P}, t) := \Tilde{H}_\theta(\hat{P}, f(t))$ where $f(t)$ enhances the model by encoding the time variable on the unit circle, representing it with two components: 
\[
f(t) := \left(\cos\left(\frac{2 \pi}{T} t\right), \sin\left(\frac{2 \pi}{T} t\right)\right)
\]
By explicitly encoding the periodicity as an inductive bias, we ensure that the resulting motion inherently satisfies the constraints of periodic behavior. As shown in the numerical experiments, this approach helps to accurately model periodic phenomena 
by embedding domain-specific knowledge directly into the learning process. Specifically, the approach effectively addresses challenges in representing complex temporal patterns, such as the shrinkage/growing movement of a toy sphere or the expansion and contraction of the aortic wall, as shown in Figure \ref{general_workflow}.

\subsection{Loss Function}
Using $H_\theta$ as parameterization of the VVF in Equation \eqref{eq:ode}, we can in principle optimize the neural network parameters $\theta$ via the loss function presented in Equation \eqref{eq:first_loss}. However, our data has the inductive bias that it is periodic in time. Although our INR parameterization of the velocity vector field is periodic in time, this does not mean the whole motion is periodic. Consequently, to ensure our network approximately represents periodic motions, such as cardiac vessel movements that repeat each cardiac cycle, we propose a periodic regularisation term, namely $R_{cycle}$. Specifically, the cycle consistency loss is defined as follows:

\[
R_{\text{cycle}} = \left( \frac{1}{n} \sum_{i=1}^{n} \left \lVert \text{P}_{0,i} - \phi_T(\text{P}_{0,i}) \right \rVert_2^2 \right)
\]
where $n$ is the number of sampled points and \(\text{P}_{0,i}\) are the point coordinates at the starting position. In summary, the INR is trained to optimize the following objective function:

\begin{equation}
    \begin{aligned}
        \min_{\theta} \quad & \left[\sum_{i=0}^{N-2} \left[ \lVert I_{t_i} \circ \phi_{t_i \to T} - I_{T}  \rVert_2^2 \right] \right] + \lambda R_{cycle}, \\
        \textrm{s.t.} \quad &   \frac{\partial \phi_t(\hat{P})}{\partial t} = H_\theta(\phi_t(\hat{P}), t), \, \phi_0(\hat{P}) = \hat{P}, \, \phi_{t \rightarrow T}:=\phi_T \circ \phi_t^{-1}
    \end{aligned}
    \label{eq:final_loss}
\end{equation}
where $\lambda \in \mathbb{R}$ is a regularization constant. 

\subsection{Data}
The model performance was evaluated using three datasets. First, three synthetic datasets were created. Each dataset consisted of a series of 25 3D segmentation images of a sphere that was deformed according to a known pattern: linear growth, exponential growth, or periodic (sinusoidal) growth and shrinkage (Figure \ref{synthetic_data_and_results}a). For each image, a corresponding triangular mesh was provided. It's worth noting that the datasets with linear and exponential growth were used at an early stage to evaluate the operation of diffeomorphic registration. In addition, since they are not periodic, the regularization term $R_{cycle}$ was not applied. Then, five 4D flow MRI datasets were included of three male (15 - 55y) and two female (77 - 79y) patients.
These datasets were acquired using a 1.5T Optim MR450w scanner (GE Medical Systems). For all datasets, manual segmentation of the end-diastolic and peak-systolic frame was performed using 3D Slicer software \cite{kikinis20133d} and a corresponding triangular mesh was obtained. Reconstructed images had an in-plane resolution of $1.4\times  1.4$ mm$^2$, and a slice thickness of $2.8$ mm. Similar to the synthetic data, each cardiac cycle consisted of 25 images $I_t$. 
Finally, three ECG-gated CT data sets including two male (43 - 80y) and one female (80y) were included to assess the method's performance in alternative modalities. These datasets were acquired with a SOMATOM Force scanner (Siemens Healthineers) with an in-plane resolution of $0.26 - 0.38\times  0.26 - 0.38$ mm$^2$, and a slice thickness of $0.69 - 1$ mm. For this specific data, each cardiac cycle consisted of 20 frames.







\subsection{Evaluation Metrics}
To assess the network performances, temporal image registration was performed on different types of data. To assess deformation quality, we used the Hausdorff Distance (HSD) to compute the distance between meshes and Peak Signal-to-Noise Ratio (PSNR) to assess image similarity. A qualitative evaluation was also carried out comparing different point trajectories in different experiments (Fig. \ref{trajectories_point}).

\section{Experiments and Results}

\subsection{Implementation Details}
All models were implemented in PyTorch. The network model \(\tilde{H}_{\theta}\) was an MLP with sinusoidal activation functions \cite{sitzmann2020implicit}, accounting for 3 layers with 256 neurons each. The \(\omega\) hyperparameter, which controls the period of the sinusoidal functions, was set to 6 for the synthetic dataset and 30 for the 4D flow MRI and CT data. 
To solve the resulting ODE appearing in optimization problem \eqref{eq:final_loss}, we use the Euler method with $N-1$ timesteps, as implemented in the Torchdiffeq library \cite{torchdiffeq}. We used Adam \cite{kingma2014adam} with a learning rate of 3e-5. Model optimization was conducted for 1000 epochs using 5000 (synthetic dataset) and 10000 (real dataset) randomly sampled points in the spatial domain.
For each sequence of images \( \{ I_{t_i} \}_{i=0,\ldots,N-1} \), the sampled spatial coordinates $\hat{P}$ were normalized to the \([-1, 1]^3\) range. Additionally, the time points \( t_0, t_1, \ldots, t_{N-1} \) were equally spaced and normalized to the \([0, 1]\) interval. Finally, in the numerical experiments below, we get a sequence of spheres or aortic walls by deforming an initial mesh $\mathcal{M}_{t_0}$ via $\phi_t^{-1}\left(\mathcal{M}_{t_0}\right)$ with $\phi_t$ the DVF in Equation \eqref{eq:final_loss}. The time required for optimization was between 12 min for 5000 points and 25 min for 10000 points for each patient.


\subsection{Synthetic Data}

\begin{figure}[t]
\includegraphics[width=\textwidth]{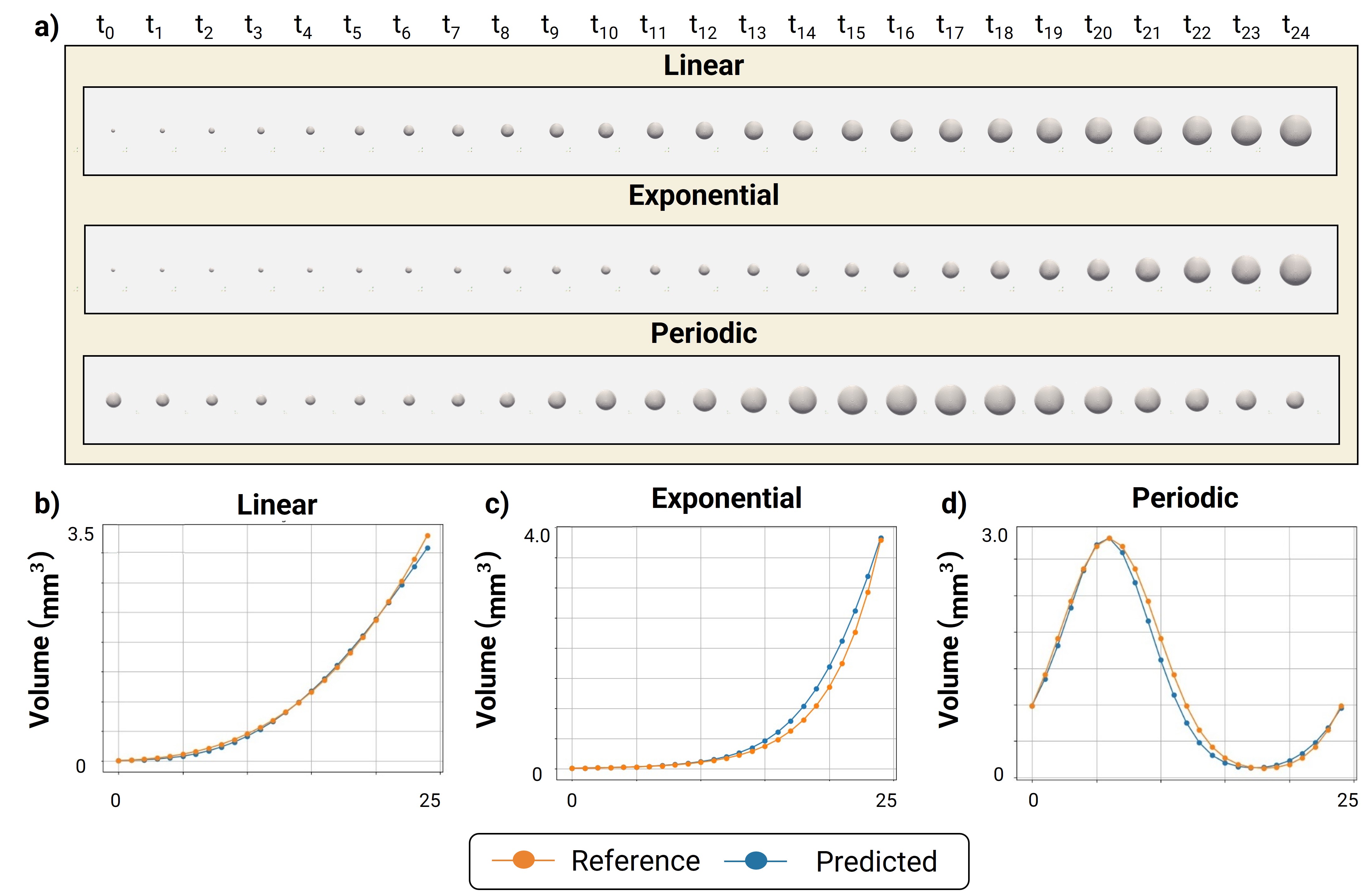}
\caption{Synthetic dataset of linear, exponential, and periodic growth patterns (a). Volume comparison of the sphere datasets for linear growth (b), exponential growth (c), and periodic growth (d) between predicted (orange) and reference (blue) lines.} \label{synthetic_data_and_results}
\end{figure}

\begin{figure}[t]
\includegraphics[width=\textwidth]{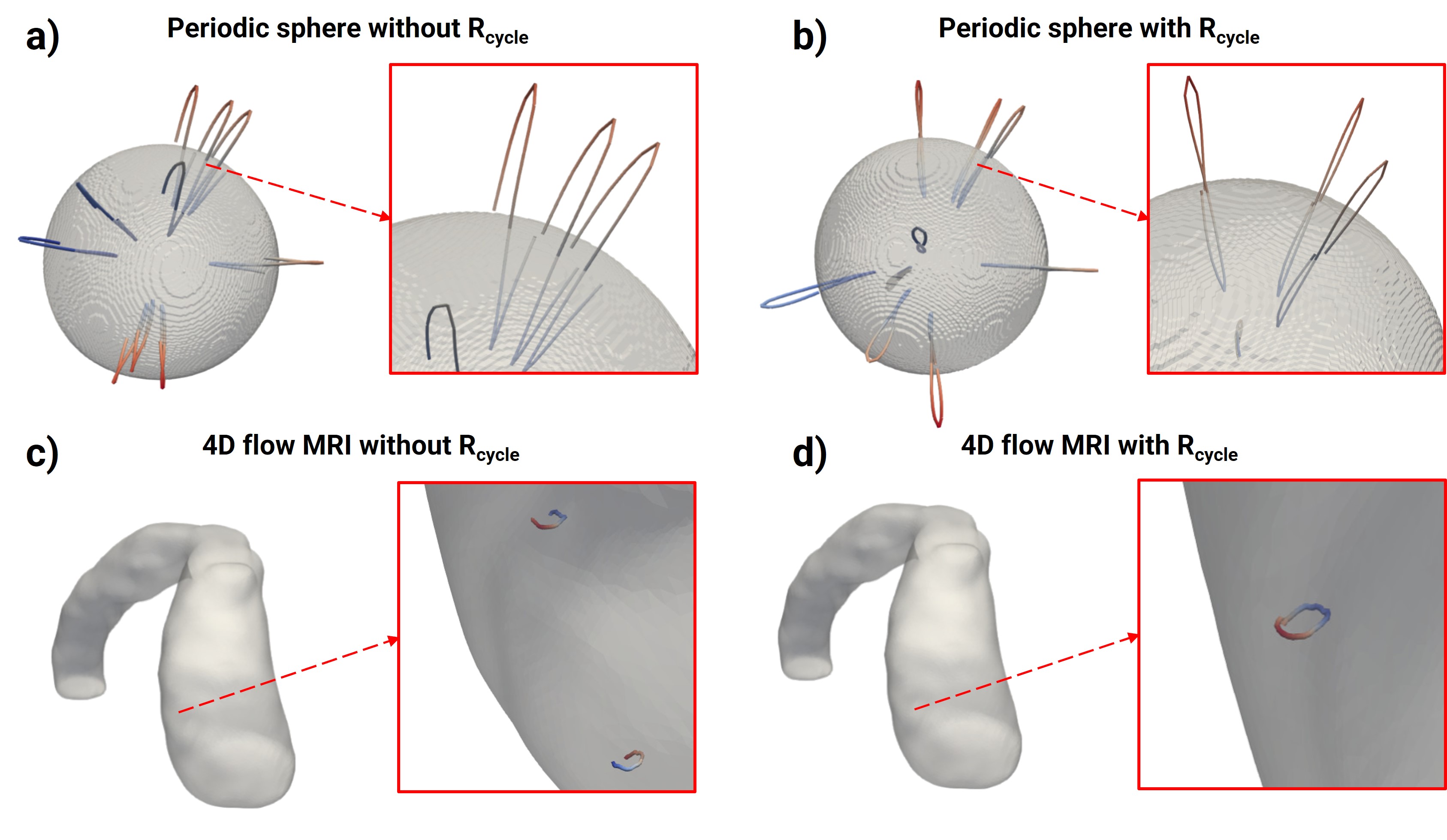}
\caption{Point trajectories tracking target deformations for periodic sphere (a, b) and real 4D flow MRI data (c, d). For both datasets, the results without $R_{cycle}$ regularisation (a, c) and with $R_{cycle}$ (b, d) are shown. $R_{cycle}$ is required to enforce periodic motion.} \label{trajectories_point}
\end{figure}

For each of the three synthetic datasets, an INR was optimized across all time points. 
For each deformed sphere surface, the volume is calculated and then compared to the ground-truth one.
The three plots in Figure \ref{synthetic_data_and_results}(b,c,d) show a comparison of the volumes of the ground truth and extracted spheres for the benchmark experiments across the linear, exponential, and periodic growth patterns respectively. Table \ref{tab:hsd-table} lists the average HSD value between predicted and reference meshes over the full cardiac cycle.
We qualitatively assessed the effect of periodic regularization in an ablation study. Figure \ref{trajectories_point} shows the sphere in combination with local point trajectories through the cardiac cycle. From this, it can be seen that regularization enforces periodic motion patterns.


\setlength{\tabcolsep}{0.5em} 
{\renewcommand{\arraystretch}{1.2} 
    \begin{table}[t!]
    \centering
    \caption{Quantitative results. The HSD and PSNR values are reported in columns for both synthetic and real data. The best metric values are reported in bold. The difference between optimization with and without temporal regularization ($R_{cycle}$) is listed.}
    \begin{tabular}{ll|cccc}
    \hline
        \toprule 
        & & \multicolumn{2}{c}{$\downarrow$ \textbf{HSD (mm)}} & \multicolumn{2}{c}{$\uparrow$ \textbf{PSNR (dB)}}\\
        & & $noR_{cycle}$ & $R_{cycle}$ & $noR_{cycle}$ & $R_{cycle}$\\

        \midrule
        \textbf{Sphere} & Linear & 8.41 & no periodicity & 12.78 & no periodicity \\
        & Exponential & 5.62 & no periodicity & 11.58 & no periodicity \\
        & Periodic & 6.23 & \textbf{4.59} & 11.25 & 10.91 \\

        \midrule
        \textbf{4D flow MRI} & Sample 1 & 7.44 & \textbf{7.34} & 29.65 & 27.46 \\
        & Sample 2 & 7.34 & \textbf{6.61} & 27.04 & 26.92 \\
        & Sample 3 & 5.78 & \textbf{4.98} & 25.53 & 25.34 \\
        & Sample 4 & 5.89 & \textbf{5.55} & 26.32 & 26.03 \\
        & Sample 5  & 4.27 & \textbf{3.96} & 25.86 & 25.73 \\
        \midrule
        \textbf{CT data} & Sample 1 & 4.98 & \textbf{3.62} & 18.65 & 18.45 \\
        & Sample 2 & 5.49 & \textbf{3.49} & 17.37 & 17.29 \\
        & Sample 3 & 4.33 & \textbf{3.51} & 18.48 & 18.32 \\
    \hline
    \end{tabular}
    \label{tab:hsd-table}
    \end{table}
}

\subsection{4D Flow MRI Data}

\begin{figure}
\centering
\includegraphics[width=0.8\textwidth]{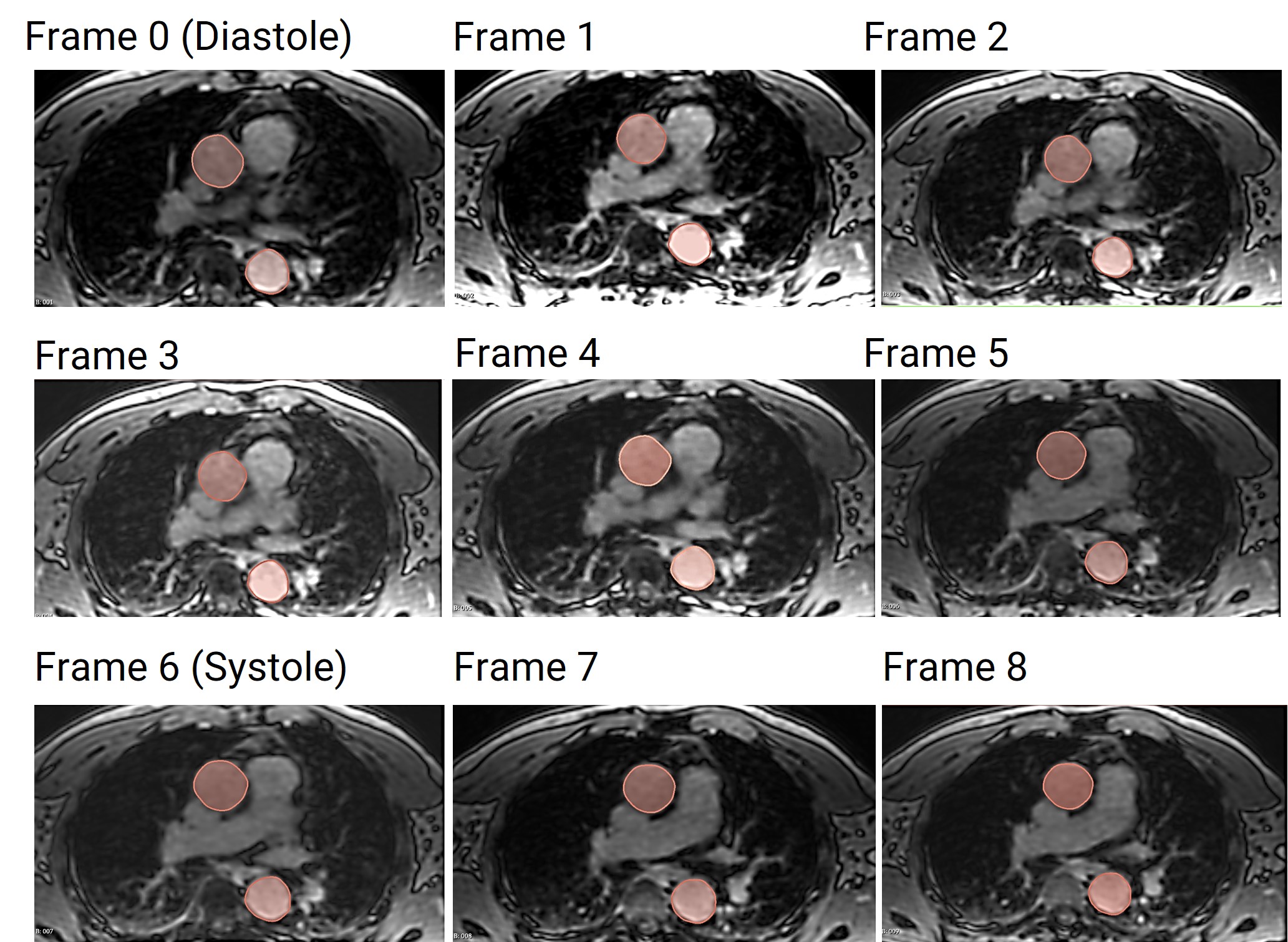}
\caption{First 9 frames} of 4D flow MRI axial slices ranging from end-diastole to peak-systole. On the same slice, the overlapping of the real target vessel in the magnitude data and the deformed first frame mesh through the estimated DVF is shown over the cardiac cycle. \label{4dflow_results}
\end{figure}

In the 4D flow MRI experiments, the INR was trained using magnitude images. Differently from the synthetic dataset, 
the 4D flow MRI datasets account for just two frame segmentations (end-diastole and peak-systole), lacking full-cycle segmentations. For this reason, the HSD was only evaluated in the frames with available meshes, with obtained results shown in Table \ref{tab:hsd-table}. The table again shows the benefit of the $R_{cycle}$ on the reconstruction performance. Additionally, Fig. \ref{trajectories_point}(c,d) shows that only with $R_{cycle}$ we obtain periodic motions.

Then, a qualitative evaluation was carried out. Figure \ref{4dflow_results} shows the results for a 4D flow MRI patient. Specifically, the deformed meshes, derived from the DVF application on the first time frame mesh, were overlapped with the first 9 corresponding ground truth time frame magnitude images. The deformed mesh overlaps well with the correct structure inside the magnitude images.

\begin{figure}
\includegraphics[width=\textwidth]{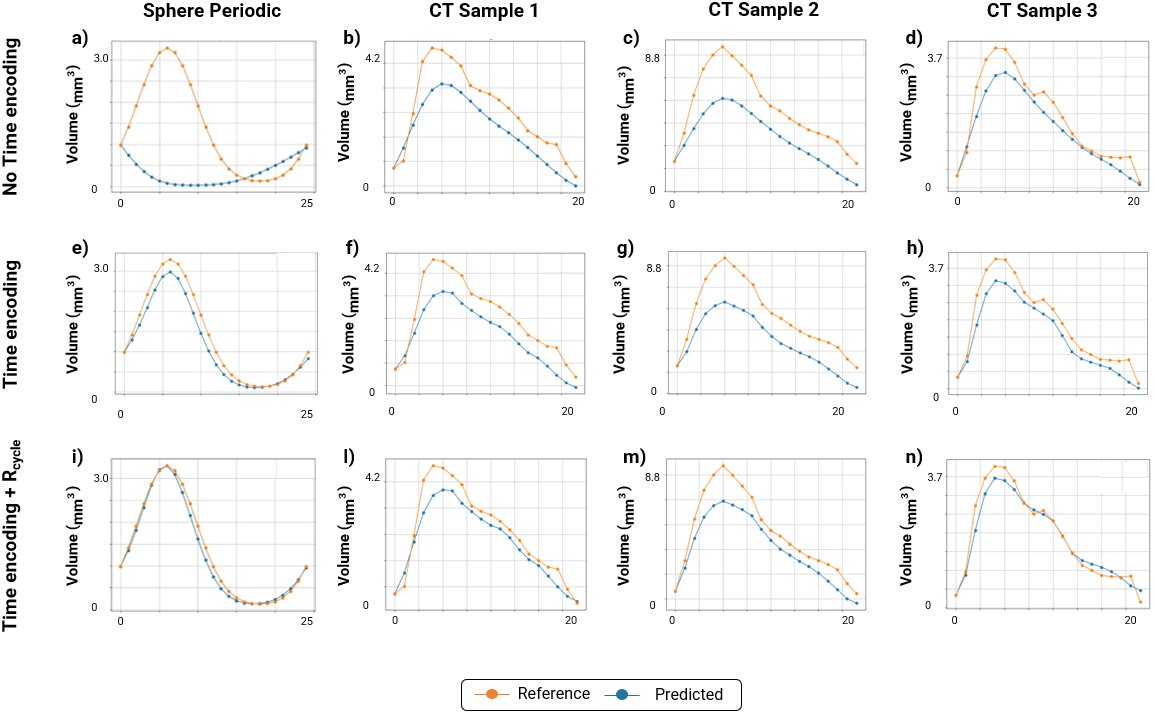}
\caption{Volume comparison of the periodic sphere (a,e,i) and CT datasets (b-d,f-h,l-n) with no time encoding (a-d), time encoding (e-h), and time encoding coupled with $R_{cycle}$ (i-n) applied.} \label{full_results}
\end{figure}

\subsection{CT Data}


For each frame in the CT data, segmentations were available. The HSD was evaluated in all frames, and the corresponding volume was evaluated and then compared to the ground truth volume. Figure \ref{full_results} shows a comparison of the volumes of the ground truth and the predicted deformed shapes. Table \ref{tab:hsd-table} lists the average HSD value between predicted and reference meshes over the full cardiac cycle. In this case also, the benefit of time encoding and $R_{cycle}$ on reconstruction performance can be appreciated from both Figure \ref{full_results} f-h,l-m and Table \ref{tab:hsd-table}.

\section{Discussion and Conclusion}
We have presented a time-dependent and periodic VVF modeling approach, continuous in both spatial and temporal domains, based on INRs to be applied to 4D cardiac data. A diffeomorphism was obtained by integrating an ODE solver into our model, 
and semi-supervised training avoided requiring large datasets. We qualitatively and quantitatively assessed the accuracy of estimated DVFs on synthetic periodic data, and different real data. 

Regarding the contribution of $R_{cycle}$, the point trajectories tracked in both synthetic (Fig. \ref{trajectories_point}b) and real (Fig. \ref{trajectories_point}d) periodic data with regularisation applied showed the model's ability to represent time-dependent and periodic motion even more accurately. This effect can also be noticed in Figure \ref{full_results} l-n and Table \ref{tab:hsd-table}. 
Regarding the effect of time encoding, Figure \ref{full_results} shows its effect on different data types. Specifically, on synthetical data with rapid periodic growth patterns (Fig. \ref{full_results} a,e), the network without time encoding was unable to correctly represent the deformation, both in time and space. Regarding periodic cardiac data with minor abrupt deformation (Fig. \ref{full_results} b, c, d, f, g, h), the adoption of time encoding helps to obtain better accuracy, following local volume changes better than the case without time encoding. Indeed, we can notice that the volume evolution is more linear and less realistic in cases without time encoding and more locally refined in the ones that include it. Since the main objective was to build a network with a broad application of cardiac data, time encoding is a fundamental assumption for cases with different temporal deformation patterns.
Therefore, the ability of the proposed model to generalize results on different periodic data ensuring continuous representation in space and time but also time-dependency and periodicity, makes it a promising solution for clinical applications.

Even if accurate results are obtained for synthetic full-time sequences, some limitations arise with more complex real data. 
Specifically, 
preliminary experiments not shown in the paper demonstrate that, although the volume progression of the aorta during the cardiac cycle follows the correct systole-diastole pattern, a progressive reduction in VVF accuracy arises when the number of time frames shown to the networks increases. In particular, the VVF loses strength and the target wall progressively deforms less and less accurately in relation to more time frames. In future work, this might be addressed by adding additional loss terms that quantify, e.g., similarities between neighboring frames, instead of only the difference with respect to the first frame, as we have now done. Additionally, future work would include an evaluation of the proposed method on different cardiac data employing different quantitative metrics.

Our method requires the optimization of a neural network per imaging sequence. The execution time for this primarily depends on the number of points sampled per iteration and settings of the ODE solver, e.g., the use of the adjoint to limit memory usage. Moreover, we found that the complexity of the data affects the time required for optimization. Alternative approaches such as segmentation-based motion estimation methods \cite{bustamante2023automatic} or non-neural LDDMM implementations~\cite{beg2005computing} might offer a different trade-off between speed and accuracy and will be compared against in future work.



In summary, we have effectively addressed the limitations of current 4D flow MRI hemodynamic analysis methods, which traditionally assume static artery walls, by leveraging neural fields to estimate continuous time-dependent periodic wall deformations throughout the cardiac cycle. Our method allows us to model time-dependent velocity fields facilitating the deformation of segmentation masks or meshes over time in a Lagrangian framework, enabling detailed visualization and quantification of local wall motion patterns. 
The periodic regularization $R_{cycle}$ ensures that the periodic nature of cardiovascular data is appropriately captured, enhancing the point trajectories over time. This strategy also helps the network in increasing the DVF estimation accuracy on full-cycle cardiac sequences (Fig. \ref{full_results}).
Overall, our neural fields-based method improves the tracking analysis of dynamic vascular structures, paving the way for more accurate and comprehensive hemodynamic assessments in cardiovascular research and clinical practice.

%
%
%
\bibliography{references}

\end{document}